\RequirePackage{fix-cm}
\documentclass[twocolumn]{svjour3}          
\smartqed  
\usepackage{graphicx}
\usepackage{latexsym}
\usepackage{times}
\usepackage{epsfig}
\usepackage{graphicx}
\usepackage{amsmath}
\usepackage{amssymb}

\usepackage{graphicx}
\usepackage{amsmath}
\usepackage{amssymb}
\usepackage{booktabs}

\usepackage{subfig}
\usepackage{algorithm}
\usepackage{algorithmic}
\usepackage{multirow}
\usepackage[table,xcdraw]{xcolor}
\usepackage{booktabs}
\usepackage{amsmath}
\usepackage{bbding}
\usepackage{enumitem}

\usepackage{multirow}
\usepackage[normalem]{ulem}
\useunder{\uline}{\ul}{}
\usepackage{url}            
\usepackage[pagebackref=true,breaklinks=true,letterpaper=true,colorlinks,bookmarks=false]{hyperref}

\emergencystretch=\maxdimen
\begin{document}

\title{DeepFake-Adapter: Dual-Level Adapter for DeepFake Detection}


\author{Rui Shao         \and
        Tianxing Wu \and
        Liqiang Nie \and
        Ziwei Liu
}


\institute{Rui Shao\at
	School of Computer Science and Technology, Harbin Institute of Technology (Shenzhen)  \\
     shaorui@hit.edu.cn   
           \and
           Tianxing Wu\at 
           S-Lab, Nanyang Technological University \\
         twu012@ntu.edu.sg
         \and
       Liqiang Nie\at
	School of Computer Science and Technology, Harbin Institute of Technology (Shenzhen)  \\
      nieliqiang@gmail.com   
           \and
        Ziwei Liu \at 
           S-Lab, Nanyang Technological University \\
         ziwei.liu@ntu.edu. \\
         \\
    Ziwei Liu is the corresponding author.
}

\date{Received: date / Accepted: date}

\maketitle

\begin{abstract}
Existing deepfake detection methods fail to generalize well to unseen or degraded samples, which can be attributed to the over-fitting of low-level forgery patterns. Here we argue that high-level semantics are also indispensable recipes for generalizable forgery detection. Recently, large pre-trained Vision Transformers (ViTs) have shown promising generalization capability. In this paper, we propose the first parameter-efficient tuning approach for deepfake detection, namely \textbf{DeepFake-Adapter}, to effectively and efficiently adapt the generalizable high-level semantics from large pre-trained ViTs to aid deepfake detection. Given large pre-trained models but limited deepfake data, DeepFake-Adapter introduces lightweight yet dedicated dual-level adapter modules to a ViT while keeping the model backbone frozen. Specifically, to guide the adaptation process to be aware of both global and local forgery cues of deepfake data, \textbf{1)} we not only insert \textbf{Globally-aware Bottleneck Adapters} in parallel to MLP layers of ViT, \textbf{2)} but also actively cross-attend \textbf{Locally-aware Spatial Adapters} with features from ViT. Unlike existing deepfake detection methods merely focusing on low-level forgery patterns, the forgery detection process of our model can be regularized by generalizable high-level semantics from a pre-trained ViT and adapted by global and local low-level forgeries of deepfake data. Extensive experiments on several standard deepfake detection benchmarks validate the effectiveness of our approach. Notably, DeepFake-Adapter demonstrates a convincing advantage under cross-dataset and cross-manipulation settings. The code has been released at \href{https://github.com/rshaojimmy/DeepFake-Adapter/} {https://github.com/rshaojimmy/DeepFake-Adapter/}.

\end{abstract}

\section{Introduction}
With recent advances in deep generative models, increasing hyper-realistic face images or videos can be readily generated, which can easily cheat human eyes. This leads to serious misinformation and fabrication problems in politics~\cite{shao2023detecting,shao2024detecting,shao2022detecting,shao2022open,shao2023robust}, entertainment and society once these techniques are maliciously abused. This threat is known as \textit{DeepFake}.

\begin{figure}[t] 
\centering
	\subfloat[FaceForensics++ DeepFakes]{\includegraphics[width = 0.48\textwidth]{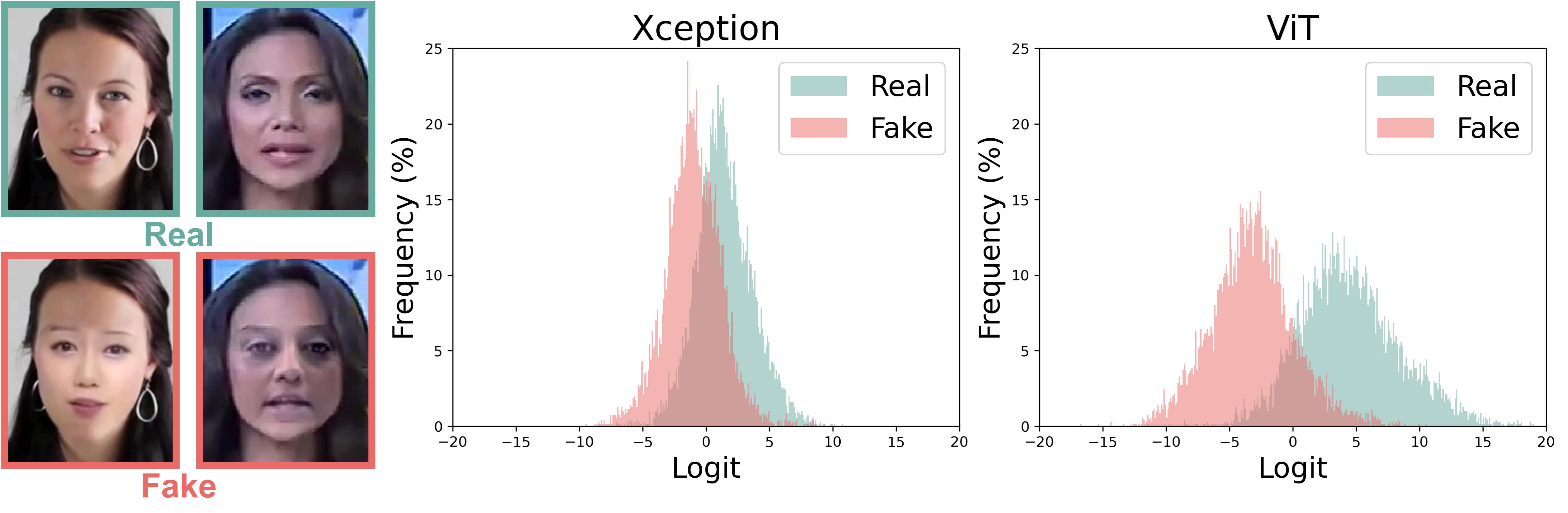}}
	\hfill
	\subfloat[FaceForensics++ FaceSwap]{\includegraphics[width = 0.48\textwidth]{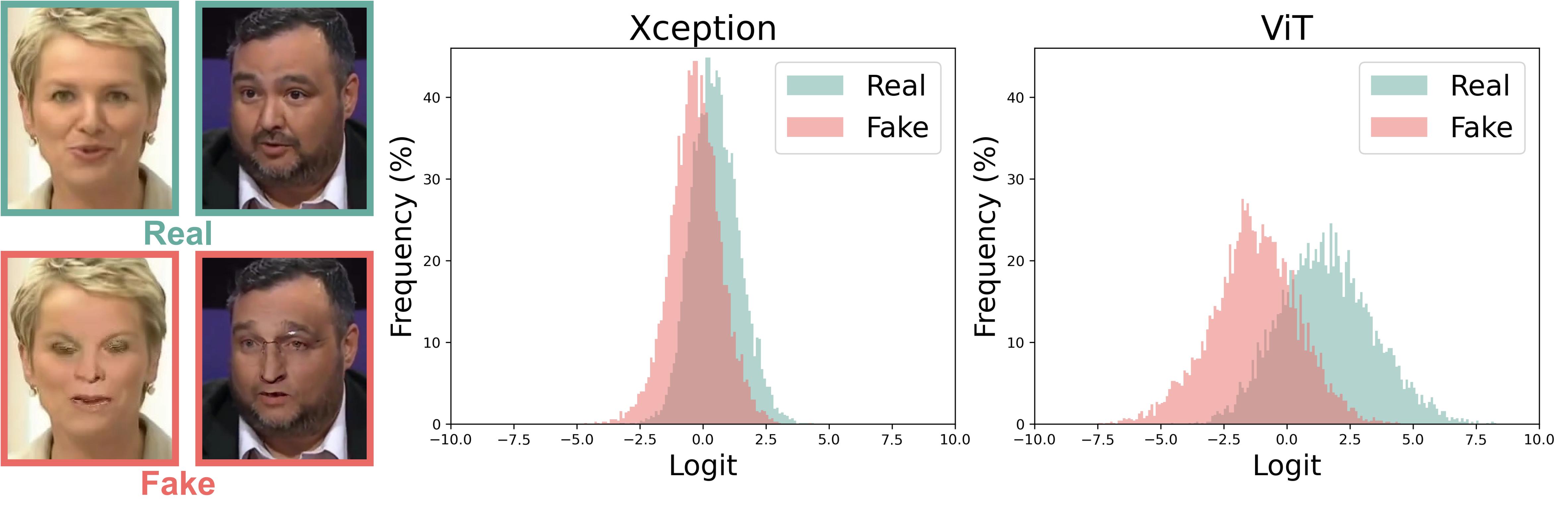}}
\caption{Example images and distributions of pre-trained Xception and ViT features after linear-probe on (a) DeepFake and (b) FaceSwap splits of FaceForensics++ dataset.}
\label{fig:intro}
\end{figure}

To address this security issue, various deepfake detection methods have been proposed and obtain promising performance when training and testing forgery data are from identical manipulation types with good quality. Nevertheless, their performance degrades once countering unseen or low-quality forgeries~\cite{luo2021generalizing,Patch-based,shao2022federated,shao2020regularized}. This may be because most of existing deepfake detection methods merely focus on exploiting low-level forgery features from local textures~\cite{chen2021local,gu2022delving,zhao2021multi,liu2020global,shao2019multi,shao2018joint}, blending boundary~\cite{li2020face}, or frequency information~\cite{li2021frequency,qian2020thinking}. These features have the following commonness in practice: \textbf{1)} different forgeries tend to have quite distinct low-level characteristics and thus testing data with unseen forgery types would present quite distinct forgery patterns compared to training data and \textbf{2)} a portion of low-level forgery patterns are likely to be altered and covered by post-processing steps such as compression, blur and noise in low-quality data. These factors degenerate the generalization ability of extracted forgery representations. To address these issues, this paper explores high-level semantics to facilitate a generalizable deepfake detection. In particular, as shown in example images of Fig.~\ref{fig:intro}, we can observe that apart from distinct low-level patterns such as textures between real and fake faces, some generic high-level semantics of real faces such as face style and shape are also altered by some face manipulation methods (\textit{e.g.,} DeepFake and FaceSwap in FaceForensics++ dataset~\cite{rossler2019faceforensics++}). Thus, these high-level semantics could be exploited for deepfake detection as they are robust to variation of low-level features.

Recently, Vision Transformer (ViT)~\cite{dosovitskiy2020image} and its variants have demonstrated remarkable success in a broad range of computer vision tasks. Various large ViTs pre-trained on massive labeled data are able to learn representations with rich semantics. We preliminarily verify the efficacy of high-level semantic features of large pre-trained ViT for deepfake detection in Fig.~\ref{fig:intro}. Following the setting of linear-probe evaluation in~\cite{radford2021learning}, we compare the linear separability regarding real/fake faces of FaceForensics++  dataset~\cite{rossler2019faceforensics++} based on features extracted by Xception~\cite{chollet2017xception} pre-trained on ImageNet-1K~\cite{deng2009imagenet} and ViT-Base~\cite{dosovitskiy2020image} pre-trained on ImageNet-21K, respectively. As illustrated in Fig.~\ref{fig:intro}, high-level semantic features from both pre-trained Xception and ViT have the potential to discriminate face forgeries through a simple linear-probe. Furthermore, the separation between real/fake distributions of ViT features are substantially larger than that of Xception features on both manipulation types. These observations demonstrate that \textbf{1)} high-level semantic features are useful for deepfake detection and \textbf{2)} features from larger pre-trained ViT model are more effective for deepfake detection. This motivates us to dig the power of large pre-trained ViTs for our task.

A straightforward way to adapt the pre-trained ViT for deepfake detection is full fine-tuning (full-tuning) with face forgery data. However, the performance of full-tuning would be severely affected by two factors \textbf{1)} given a large volume of pre-trained ViT's parameters (\textit{e.g.,} ViT-Base~\cite{dosovitskiy2020image} with 85.8M parameters) and a relatively smaller amount of deepfake detection data, full-tuning is very likely to result in over-fitting and thus damages the generalization ability of ViT and \textbf{2)} as proved in~\cite{kumar2022fine}, full-tuning could distort pre-trained features and leads to worse performance in the presence of large distribution shift. Therefore, to effectively and efficiently adapt generalizable high-level semantics from large pre-trained ViTs to deepfake detection, this paper proposes to explore a fast adaptation approach, namely \textbf{DeepFake-Adapter}, in a parameter-efficient tuning manner. DeepFake-Adapter allows a small amount (16.9M, 19\% of ViT-Base parameters. Refers to Table~\ref{tab:config}) of model parameters, \textit{i.e.,} adapter modules, to be trained whereas the vast majority of pre-trained parameters in the model backbone are kept frozen. 

Notably, DeepFake-Adapter consists of dual-level modules. First, to adapt ViT with global low-level features, we insert \textbf{Globally-aware Bottleneck Adapters (GBA)} in parallel to Multilayer Perception (MLP) layers of the pre-trained ViT. It explores global low-level forgeries, \textit{e.g.,} blending boundary~\cite{li2020face}, in a bottleneck structure. Second, to capture more local low-level forgeries in the adaption process, \textit{e.g.,} local textures~\cite{chen2021local,gu2022delving,zhao2021multi,liu2020global}, we devise \textbf{Locally-aware Spatial Adapters (LSA)} to extract local low-level features and lead them to interact with features from the pre-trained ViT via a series of cross-attention. In this way, the forgery detection is regularized by generalizable high-level semantics from a pre-trained ViT and adapted with global and local low-level forgeries by the dual-level adapter. This organic interaction between high-level semantics and global/local low-level forgeries contributes to better generalizable forgery representations for deepfake detection. Main contributions of our paper are:

\begin{itemize}[leftmargin=*]
\item We argue that high-level semantics of large pre-trained ViTs could be beneficial for deepfake detection. To make use of these semantics, we are the first work to introduce the \textit{adapter} technique into the field of deepfake detection, which fast adapts a pre-trained ViT for our task.
\item We propose a novel \textbf{DeepFake-Adapter}, which is a dual-level adapter composed of \textbf{Globally-aware Bottleneck Adapters (GBA)} and \textbf{Locally-aware Spatial Adapters (LSA)}. \textbf{DeepFake-Adapter} can effectively adapt a pre-trained ViT by enabling high-level semantics from ViT to organically interact with global and local low-level forgeries from adapters. This contributes to more generalizable forgery representations for deepfake detection.
\item Extensive quantitative and qualitative experiments demonstrate the superiority of our method for deepfake detection. Notably, DeepFake-Adapter outperforms the full-tuning adaptation method by only tuning less than 20\% of all model parameters.
We hope that our approach can facilitate future research on generalizable deepfake detection in the era of larger vision models.
\end{itemize}

\begin{figure*}[t] 
	\begin{center}
		\includegraphics[width=1\linewidth]{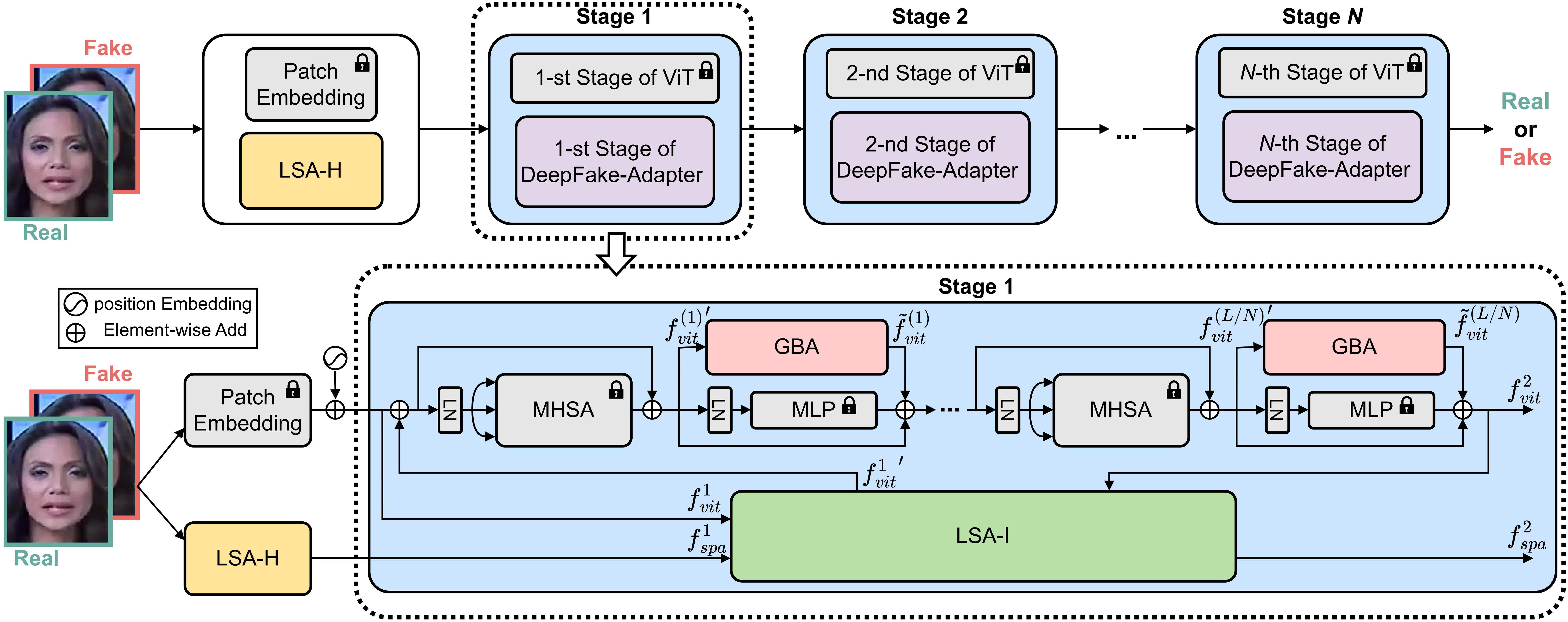}
	\end{center}
	\caption{\textcolor{black}{Overall architecture of proposed model. The model consists of $N$ stages. Each stage contains MHSA and MLP layers of pre-trained ViT, GBA and LSA (LSA-H and LSA-I) of proposed DeepFake-Adapter.}}
	\label{fig:overview}
\end{figure*}

\section{Related Work}
\subsection{DeepFake Detection}
Current deepfake detection methods can be mainly categorized into spatial-based and frequency-based forgery detection. The majority of spatial-based deepfake detection methods pay attention to capturing low-level visual cues from the spatial domain. The blending boundary caused by face forgery operations is detected as the visual artifacts for deepfake detection~\cite{li2020face}. Various local textures~\cite{chen2021local,gu2022delving,zhao2021multi,zhu2021face,liu2020global} are intensively analyzed and explored to highlight the appearance differences between real and forged faces. Besides, direct light~\cite{zhu2021face} is disentangled by a 3D decomposition method and fused with other features using a two-stream network for forgery detection. Patch diffusion~\cite{zhang2022patch} and patch inconsistency~\cite{zhao2021learning} are also studied to explore the distinct correlation consistency among local patch features between real and forgery faces. Moreover, motion artifacts are dig out from mouth movements as the face forgery patterns by fine-tuning a temporal network pretrained on lipreading~\cite{haliassos2021lips}. \textcolor{black}{This method targets at detecting fake videos based on mouth movements without overfitting to low-level, manipulation-specific artefacts. RealForensics~\cite{haliassos2022leveraging} is also another work to exploit generalizable high-level temporal features by studying the natural correspondence between the visual and auditory modalities based on a self-supervised cross-modal manner.} In addition, noise characteristics are exploited as the forgery clues in works~\cite{gu2022exploiting,zhou2017two}.

On the other hand, some methods focus on frequency domain for detecting spectrum artifacts. High-frequency part of Discrete Fourier Transform (DFT)~\cite{durall2019unmasking,dzanic2020fourier} are extracted to detect distinct spectrum distributions and characteristics between real and fake images. Local frequency statistics based on Discrete Cosine Transform (DCT) are exploited by F$^{3}$-Net~\cite{qian2020thinking} to mine forgery cues. Up-sampling artifacts in phase spectrum are explored by a Spatial-Phase Shallow Learning method~\cite{liu2021spatial}. To capture generalizable forgery features, high-frequency features are integrated with regular RGB features with a two-stream model~\cite{luo2021generalizing}. What's more, a frequency-aware discriminative feature learning framework is proposed to perform metric learning in frequency features~\cite{li2021frequency}.

Most of the above deepfake detection methods only study low-level spatial or frequency artifacts. Instead, this paper performs interaction between high-level semantics from a large pre-trained ViT and dual levels of forgeries from DeepFake-Adapter, unveiling better generalizable forgery representations.
\vspace*{-5mm}
\subsection{Parameter-Efficient Transfer Learning}
Parameter-efficient tuning methods have drawn increasing attention starting from the natural language processing (NLP) community. Unlike most of the popular transfer learning methods such as full-tuning and linear-probe~\cite{zhuang2020comprehensive}, parameter-efficient tuning methods only need to train a small portion of model parameters in consideration of the rapid increase in model size of large pre-trained language models~\cite{li2024optimus,chen2024lion,shen2024mome,ye2024cat}. Prompt learning~\cite{liu2021pre,lester2021power} is wide-used in NLP which prepends learnable tokens into the input text. Adapter~\cite{houlsby2019parameter} and LoRA~\cite{hu2021lora} add tiny learnable modules into NLP transformers. Some follow-up parameter-efficient tuning works in the computer vision field~\cite{jia2022vpt,chen2022adaptformer,chen2022vision,chen2022vitadapter} have also been proposed recently. This paper is the first work to introduce the \textit{adapter} technique into deepfake detection with a dedicated dual-level adapter.
\section{Our Approach}
\subsection{Overview}
The overall architecture of the proposed network is illustrated in Fig.~\ref{fig:overview}. As depicted in the first row of Fig.~\ref{fig:overview}, the whole network is composed of $N$ stages. Each stage contains one stage of pre-trained ViT whose parameters are frozen during the training, and one stage of deepfake-adapter with trainable parameters for fast adaptation. Moreover, the patch embedding layer of ViT is also frozen and the head part of the proposed Locally-aware Spatial Adapter (LSA-H) is inserted at the beginning of the network.

Specifically, we take Stage 1 as an example shown in the second row of Fig.~\ref{fig:overview}. Given a pre-trained ViT with total $L$ blocks (each block consists of a Multi-Head Self-Attention (MHSA) layer and a MLP layer), we evenly group these blocks into $N$ stages and thus the above one stage of ViT contains $L/N$ blocks. The corresponding stage of DeepFake-Adapter consists of $L/N$ Globally-aware Bottleneck Adapters (GBA) and one interaction part of Locally-aware Spatial Adapter (LSA-I), which are organically interacted with ViT for adaptation. Details of each module in the whole model are introduced in the following sections.

\begin{figure*}[t] 
	\begin{center}
		\includegraphics[width=1\linewidth]{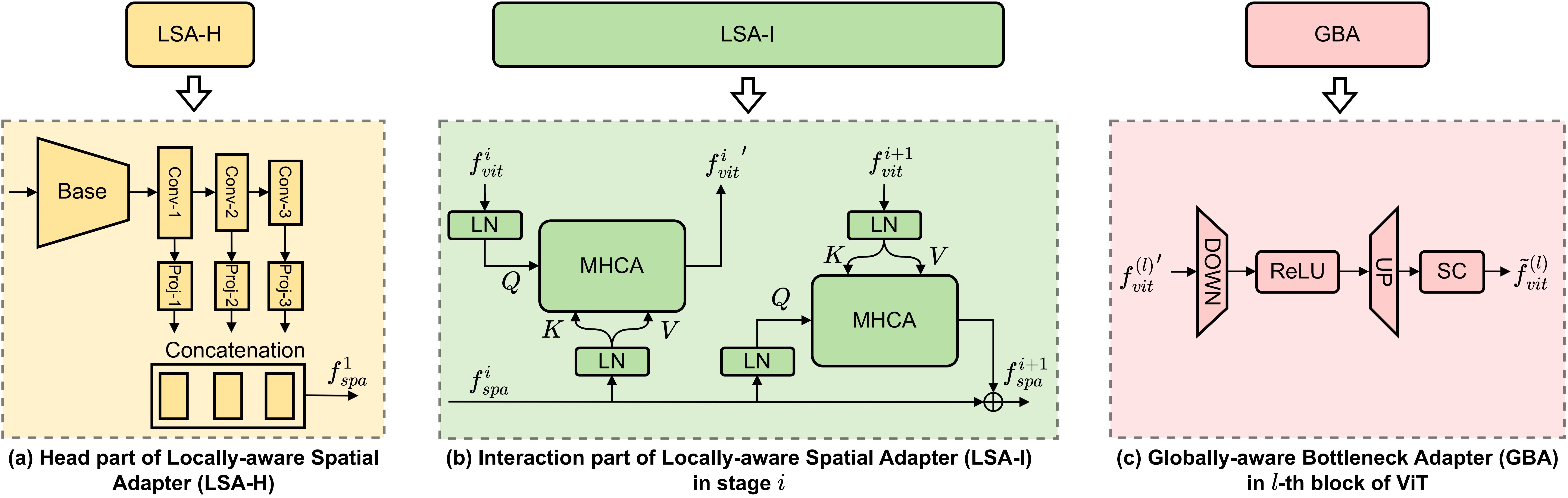}
	\end{center}
	\caption{\textcolor{black}{Details of GBA and LSA of proposed DeepFake-Adapter. (a) Head part and (b) Interaction part of LSA capture local low-level forgeries that interact with features from pre-trained ViT via a series of cross-attention. (c) GBA adapts the pre-trained ViT with global low-level forgeries in a bottleneck structure.}}
	\label{fig:submodule}
\end{figure*}

\subsection{Vanilla ViT}
We adopt a pre-trained vanilla ViT as the frozen backbone of our network. As mentioned above, it basically consists of a patch embedding layer and following $L$ blocks. Given an input image $x$ $\in$ R$^{3 \times H \times W}$, we feed it into the patch embedding layer of ViT. It firstly divides the image into non-overlapping $P \times P $ patches and then flattens them into sequential patches $x_p$ $\in$ R$^{K \times (P^2C)}$, where $(H, W)$ is the resolution of the input image; $C$ is the number of channels; $(P, P)$ is the resolution of each image patch, and $K = HW/P^2$  is the resulted number of patches. All of these image patches are projected to $D$-dimensional embedding and added with position embedding. This produces the patch embedding of ViT as the input of Stage 1, denoted as $f_{vit}^1$ $\in$ R$^{(P^2C)\times D}$. After that, the patch embedding passes through MHSA layers and MLP layers in every block of ViT to carry out a series of self-attention. Specifically, MHSA in $l$-th block of ViT is performed on normalized query ($Q$), key ($K$), and value ($V$) features as follows,
\begin{equation}
\begin{split}
{f_{vit}^{(l)}}' & = \text{Attention}(Q=\widehat{f_{vit}^{(l)}}, K=\widehat{f_{vit}^{(l)}}, V=\widehat{f_{vit}^{(l)}}) \\
&=\text{Softmax}(K^T Q / \sqrt{D})V
\end{split}
\end{equation}
where $\widehat{f_{vit}^{(l)}} = \text{LN}(f_{vit}^{(l)})$, which is the feature normalized by the LayerNorm layer~\cite{ba2016layer} as the input of MHSA layer in $l$-th block. Its output ${f_{vit}^{(l)}}'$ is then fed into the following MLP layer as follows,
\begin{equation}
\begin{split}
f_{vit}^{(l+1)} = \text{MLP}(\widehat{{f_{vit}^{(l)}}'}) + {f_{vit}^{(l)}}'
\end{split}
\end{equation}
where $f_{vit}^{(l+1)}$ is the output of $l$-th block of ViT. 

\subsection{Globally-aware Bottleneck Adapter}
Considering that MHSA layers of ViT tend to extract global features~\cite{dosovitskiy2020image}, we insert Globally-aware Bottleneck Adapter (GBA) after each MHSA layer and in parallel to MLP layers of ViT as illustrated in second row of Fig.~\ref{fig:overview}. It attempts to adapt the pre-trained ViT with more global low-level forgery features, such as blending boundary~\cite{li2020face}. Specifically, inspired by~\cite{chen2022adaptformer,he2021towards}, as shown in \textcolor{black}{Fig.~\ref{fig:submodule} (c)}, GBA is devised as a bottleneck structure in purpose of saving parameters for fast adaptation, which consists of a down-projection linear layer (DOWN) and an up-projection linear layer (UP). In addition, a ReLU layer~\cite{agarap2018deep} is incorporated between two projection layers for non-linear transformation. To adaptively weigh the importance of global low-level features in the adaptation, one more learnable scale function (SC) is added after two projection layers. The whole adaptation process of GBA is as follows,
\begin{equation}
\begin{split}
\tilde{f}_{vit}^{(l)} = \text{SC} \cdot \text{UP} \cdot \text{ReLU} (\text{DOWN}({f_{vit}^{(l)}}'))
\end{split}
\end{equation}
where $\tilde{f}_{vit}^{(l)}$ is the adapted global low-level features from corresponding GBA in $l$-th block of ViT, which can be further fused with the original output of MLP layer as follows,
\begin{equation}
\begin{split}
f_{vit}^{(l+1)} = \text{MLP}(\widehat{{f_{vit}^{(l)}}'}) + {f_{vit}^{(l)}}' + \tilde{f}_{vit}^{(l)}
\end{split}
\label{equ:global_interaction}
\end{equation}

\begin{table*}[t]
\footnotesize
\centering
\caption{Configurations of pre-trained ViT and proposed Deepfake-Adapter.}
\begin{tabular}{ccccc|ccc|ccc|c}
\toprule[1.1pt]
\multicolumn{5}{c|}{Settings of ViT~\cite{dosovitskiy2020image}}    & \multicolumn{3}{c|}{Settings of GBA} & \multicolumn{3}{c|}{Settings of LSA} & \multirow{2}{*}{\begin{tabular}[c]{@{}c@{}}Total\\ Param\end{tabular}} \\
Blocks & Width & MLP  & Heads & \#Param & N       & Width      & \#Param      & N       & Heads       & \#Param      &                                                                        \\ \hline
12     & 768   & 3072 & 12    & 85.8M   & 12      & 64          & 1.19M        & 3       & 6           & 15.73M       & 102.7M                                                                 \\ \bottomrule[1.1pt]
\end{tabular}
\label{tab:config}
\vspace*{-4mm}
\end{table*}
\subsection{Locally-aware Spatial Adapter}
It is well known that ViT has much less image-specific inductive bias, \textit{e.g.,} spatial locality, than Convolutional Neural Networks (CNNs)~\cite{dosovitskiy2020image}. This makes a ViT less likely to differentiate local low-level features between real and fake faces. To address this issue, we introduce Locally-aware Spatial Adapter (LSA) in this section, which is composed of head and interaction parts. It aims to adapt more local low-level forgery features, such as local textures~\cite{chen2021local,gu2022delving,zhao2021multi,liu2020global} for our task.
\noindent \textbf{Head part (LSA-H).} 
Inspired by recent works~\cite{yuan2021incorporating,wu2021cvt} that integrate convolutional operations of CNNs into a ViT, we introduce the convolution-based head part of LSA. It locates in parallel to the patch embedding layer of ViT, attempting to capture more local low-level forgeries of input images from the beginning. To be specific, as depicted in \textcolor{black}{Fig.~\ref{fig:submodule} (a)}, following the structure of ResNet~\cite{he2016deep}, we employ a standard CNN as the base network to extract base feature maps, which consists of three Convolution-BatchNorm-ReLU blocks and a max-pooling layer. Then, three similar convolutional blocks are used to extract several intermediate feature maps. They are composed of various pyramid resolutions, 1/$r_1$, 1/$r_2$, and 1/$r_3$ resolutions, corresponding to the size of original input images. After that, all of them are projected into the same dimension $D$ via three projectors and concatenated into a feature vector denoted as $f_{spa}^1$ $\in$ R$^{(\frac{HW}{r_1^2}+\frac{HW}{r_2^2}+\frac{HW}{r_3^2})\times D}$. Based on this, LSA-H aggregates features with diverse spatial resolutions, capturing fine-grained and rich local forgeries.
\noindent \textbf{Interaction part (LSA-I).}
Given the aggregated features $f_{spa}^1$, we intend to enable the whole adaption process to sufficiently be aware of local low-level forgeries captured from these features. To this end, as illustrated in the second row of Fig.~\ref{fig:overview}, we devise the interaction part of LSA which leads these features (\textit{e.g.,} $f_{spa}^1$ in Stage 1) to interact with features from the beginning and end of ViT in each stage (\textit{e.g.,} $f_{vit}^1$ and $f_{vit}^2$ in Stage 1). In greater detail, in Stage $i$, the first interaction is performed by a multi-head cross-attention (MHCA) between feature $f_{spa}^i$ and the feature from the beginning of ViT $f_{vit}^i$, as depicted in \textcolor{black}{Fig.~\ref{fig:submodule} (b)}. Here, we treat normalized $f_{vit}^i$ as query and normalized $f_{spa}^i$ as key and value in this MHCA as follows,
\begin{equation}
\begin{split}
{f_{vit}^{i}}' = f_{vit}^{i} + \text{Attention}(Q=\widehat{f_{vit}^i}, K=\widehat{f_{spa}^i}, V=\widehat{f_{spa}^i})
\end{split}
\label{equ:local_interaction1}
\end{equation}
where ${f_{vit}^{i}}'$ is the result of the first interaction. It will be fed back into ViT and go through the following MHSA, MLP layers and GBA modules in ViT. This adaptation process injects local low-level features into the forward process of ViT. Once obtaining the feature (denoted as $f_{vit}^{i+1}$) through the whole forward process of ViT in Stage $i$, we perform the second interaction at the end of ViT by conducting MHCA between $f_{spa}^i$ and $f_{vit}^{i+1}$. We switch the $K, Q, V$ by taking normalized $f_{spa}^i$ as query and normalized $f_{vit}^{i+1}$ as key and value in this MHCA as follows,
\begin{equation}
\begin{split}
f_{spa}^{i+1} = f_{spa}^{i} + \text{Attention}(Q=\widehat{f_{spa}^i}, K=\widehat{f_{vit}^{i+1}}, V=\widehat{f_{vit}^{i+1}})
\end{split}
\label{equ:local_interaction2}
\end{equation}
where $f_{spa}^{i+1}$ is the updated low-level features which will be forwarded to interaction with new features of ViT $f_{vit}^{i+1}$ in the next stage. As such, the influence of local low-level features regarding face forgery data could be further strengthened in the adaptation process at the end of each stage. 

\textcolor{black}{After we extract the feature $f_{spa}^{N+1}$  through $N$ stages of our model, we feed it into a classifier based on a linear layer $\rm CLS$ and calculate a cross-entropy loss as follows,
\begin{equation}
\mathcal{L}= {\rm \textbf{H}}({\rm CLS}(f_{spa}^{N+1}), y)
\end{equation}
where $y$ are labels for corresponding samples and ${\rm \textbf{H}(\cdot)}$ is the cross-entropy function. We train all the above adapters with this loss function $\mathcal{L}$ in an end-to-end manner.}

In summary, through making high-level semantic features from the pre-trained ViT interacted with global low-level features in Eq.~\ref{equ:global_interaction} and local low-level features in Eq.~\ref{equ:local_interaction1}-\ref{equ:local_interaction2}, our model based on such dual-level adaptation could exploit better generalizable forgery representations.
\subsection{Merits and Limitations} 
As aforementioned, the proposed DeepFake-Adapter performs fast adaption for a large pre-trained ViT via GBA and LSA simultaneously. This dual-level adaptation contributes to a better discriminative and generalizable deepfake detection. Furthermore, this adaptation is devised in a parameter-efficient tuning manner. We only need to train remarkably fewer parameters in GBA and LSA (less than 20\% of the original large pre-trained ViT). This makes our method easily scale up to various deepfake datasets and deployed with affordable GPU machines for training. 

On the other hand, we need to point out one main limitation of the proposed method. Since our approach is designed specifically to adapt a large pre-trained ViT, the detection process is regularized by high-level semantics from it. While its advantage facilitates a generalizable deepfake detection, it brings a negative impact that the current model is likely to be unavailable for detecting the forgery based on face reenactment (\textit{e.g.,} Face2Face and NeuralTexture in FaceForensics++ dataset). This is because these types of face manipulation only present very minor low-level forgery patterns without much modification on high-level semantics. We argue that it is impractical to address all types of deepfake manipulation in a single model. Consequently, this paper mainly focuses on detecting \textcolor{black}{one of the} most popular and the highest risky face forgery methods based on face swapping.
\section{Experiments}
\subsection{Experimental Settings}
\noindent \textbf{Datasets.}
Experiments are conducted on several existing public deepfake datasets, \textit{e.g.,} \textbf{FaceForensics++ (FF++)}~\cite{rossler2019faceforensics++}, \textbf{Celeb-DF}~\cite{li2019celeb}, \textbf{Deepfake Detection Challenge (DFDC)}~\cite{dolhansky2020deepfake}, and \textbf{DeeperForensics-1.0 (DF1.0)}~\cite{jiang2020deeperforensics}. As one of the most widely-used datasets in deepfake detection, FF++ collects 1,000 original videos, and 4000 fake videos generated by corresponding four face manipulation techniques: Deepfakes (DF)~\cite{li2020face}, Face2Face (F2F)~\cite{thies2016face2face}, FaceSwap (FS)~\cite{thies2016face2face}, and NeuralTextures (NT)~\cite{thies2019deferred}. In contrast, most of the manipulation types in Celeb-DF, DFDC, and DF1.0 datasets are based on face swapping. Considering factors mentioned in section of Merits and Limitations and the most prevailing face forgery in practice, we mainly train our model on manipulation types of DF and FS in FF++ dataset. This evaluates the generalization ability of our method on forgeries related to face swapping. Moreover, we adopt both c23 (high-quality) and c40 (low-quality) versions of FF++ data in our experiments, examining deepfake detection on forgeries with various qualities.
\begin{table}[t]
\centering
\caption{Structure details of all components in LSA-H.}
\begin{tabular}{ccc}
\toprule[1.1pt]
Layer    & Chan./Stri.  & Out.Size \\ \hline
\multicolumn{3}{c}{\textbf{Base}}   \\
\multicolumn{3}{c}{Input: image}   \\
conv0-1  & 64/2         & 112      \\
conv0-2  & 64/1         & 112      \\
conv0-3  & 64/1         & 112      \\
pool0-1  & -/2          & 56       \\ \hline
\multicolumn{3}{c}{\textbf{Conv-1}}         \\
\multicolumn{3}{c}{Input: pool0-1} \\
conv1-1  & 128/2        & 28       \\ \hline
\multicolumn{3}{c}{\textbf{Conv-2}}         \\
\multicolumn{3}{c}{Input: conv1-1} \\
conv2-1  & 256/2        & 14       \\ \hline
\multicolumn{3}{c}{\textbf{Conv-3}}         \\
\multicolumn{3}{c}{Input: conv2-1} \\
conv3-1  & 256/2        & 7        \\ \hline
\multicolumn{3}{c}{\textbf{Proj-1}}         \\
\multicolumn{3}{c}{Input: conv1-1} \\
conv4-1  & 768/1        & 28       \\ \hline
\multicolumn{3}{c}{\textbf{Proj-2}}         \\
\multicolumn{3}{c}{Input: conv2-1} \\
conv5-1  & 768/1        & 14       \\ \hline
\multicolumn{3}{c}{\textbf{Proj-3}}         \\
\multicolumn{3}{c}{Input: conv3-1} \\
conv6-1  & 768/1        & 7        \\ \bottomrule[1.1pt]
\end{tabular}
\label{tab:LSA-H}
\vspace*{-3mm}
\end{table}

\subsection{Evaluation on Discrimination Ability}
\begin{table*}[t]
\footnotesize
\begin{minipage}{.68\linewidth}
\centering
\caption{Performance of intra-dataset evaluation. Best results are in bold. Second-best results are in underline.}
\vspace*{-3mm}
\begin{tabular}{c|cccc|cccc}
\toprule[1.1pt]
                          & \multicolumn{4}{c|}{FaceForensics++ (c23)}                                                  & \multicolumn{4}{c}{FaceForensics++ (c40)}                                                                \\ \cline{2-9} 
\multirow{-2}{*}{Methods} & DF           & FS           & F2F                           & NT                            & DF             & FS             & F2F                                    & NT                            \\ \hline
ResNet-50~\cite{he2016deep}                  & 98.93        & 99.64        & 98.57                         & 95.00                         & 95.36          & 94.64          & 88.93                                  & 87.50                         \\
Xception~\cite{chollet2017xception}                  & 98.93        & 99.64        & 98.93                         & 95.00                         & 96.78          & 94.64          & 91.07                                  & 87.14                         \\
LSTM~\cite{hochreiter1997long}                      & 99.64        & 98.21        & 99.29                         & 93.93                         & 96.43          & 94.29          & 88.21                                  & 88.21                         \\
C3D~\cite{tran2014c3d}                       & 92.86        & 91.79        & 88.57                         & 89.64                         & 89.29          & 87.86          & 82.86                                  & 87.14                         \\
I3D~\cite{carreira2017quo}                       & 92.86        & 96.43        & 92.86                         & 90.36                         & 91.07          & 91.43          & 86.43                                  & 78.57                         \\
TEI~\cite{liu2020teinet}                       & 97.86        & 97.50        & 97.14                         & 94.29                         & 95.00          & 94.64          & 91.07                                  & 90.36                         \\
DSANet~\cite{wu2021dsanet}                    & 99.29        & 99.64        & 99.29                         & 95.71                         & 96.79          & 95.36          & 93.21                                  & 91.78                         \\
V4D~\cite{zhang2020v4d}                        & 99.64        & 99.64        & 99.29                         & 96.07                         & 97.86          & 95.36          & 93.57                                  & 92.50                         \\
FaceNetLSTM~\cite{sohrawardi2019poster}               & 89.00        & 90.00        & 87.00                         & -                             & -              & -              & -                                      & -                             \\
Co-motion~\cite{wang2020exposing}              & 99.10        & 98.30        & 93.25                         & 90.45                         & -              & -              & -                                      & -                             \\
DeepRhythm~\cite{qi2020deeprhythm}                & 98.70        & 97.80        & 98.90                         & -                             & -              & -              & -                                      & -                             \\
ADD-Net~\cite{zi2020wilddeepfake}                & 92.14        & 92.50        & 83.93                         & 78.21                         & 90.36          & 80.00          & 78.21                                  & 69.29                         \\
S-MIL~\cite{li2020sharp}                     & 98.57        & 99.29        & 99.29                         & 95.71                         & 96.79          & 94.64          & 91.43                                  & 88.57                         \\
S-MIL-T~\cite{li2020sharp}                    & 99.64        & 100          & {\ul 99.64}                & 94.29                         & 97.14          & 96.07          & 91.07                                  & 86.79                         \\
STIL~\cite{gu2021spatiotemporal}                      & 99.64        & 100          & 99.29                         & 95.36                         & 98.21          & 97.14          & 92.14                                  & 91.78                         \\
SIM~\cite{gu2022delving}                       & \textbf{100} & \textbf{100} & 99.29                         & {\ul 96.43}                & 99.28          & 97.86          & 95.71                                  & {\ul 94.28}                \\
\rowcolor[HTML]{E3DCDC} 
Ours                      & 99.84        & 99.76        & \cellcolor[HTML]{E3DCDC}99.33 & \cellcolor[HTML]{E3DCDC}95.97 & {\ul 99.57} & {\ul 99.00} & \cellcolor[HTML]{E3DCDC}{\ul 97.50} & \cellcolor[HTML]{E3DCDC}91.25 \\

\rowcolor[HTML]{E3DCDC} \textcolor{black}{Ours$^*$}                      & \textcolor{black}{{\ul 99.85}}        & \textcolor{black}{{\ul 99.83}}       & \cellcolor[HTML]{E3DCDC}\textcolor{black}{\textbf{99.67}} & \cellcolor[HTML]{E3DCDC}\textcolor{black}{\textbf{96.52}} & \textcolor{black}{\textbf{99.65}} & \textcolor{black}{\textbf{99.20}} & \cellcolor[HTML]{E3DCDC}\textcolor{black}{\textbf{97.61}} & \cellcolor[HTML]{E3DCDC}\textcolor{black}{\textbf{94.30}} \\
\bottomrule[1.1pt]
\end{tabular}
\label{tab:intra}
\end{minipage}%
\begin{minipage}{0.35\linewidth}
\centering
\caption{Cross-manipulation evaluation trained with DF and FS.}
\begin{tabular}{c|ccc}
\toprule[1.1pt]
Methods                      & Train                & DF                                     & FS                                     \\ \hline
Xception~\cite{chollet2017xception}                     &                      & 98.44                                  & 68.67                                  \\
RFM~\cite{wang2021representative}                          &                      & 98.80                                  & 72.69                                  \\
Add-Net~\cite{zi2020wilddeepfake}                      &                      & 98.04                                  & 68.61                                  \\
Freq-SCL~\cite{li2021frequency}                     &                      & 98.91                                  & 66.87                                  \\
MultiAtt~\cite{zhao2021multi}                     &                      & 99.51                                  & 67.33                                  \\
RECCE~\cite{cao2022end}                        &                      & \textbf{99.65}                         & {\ul 74.29}                                 \\
\cellcolor[HTML]{E3DCDC}Ours & \multirow{-7}{*}{DF} & \cellcolor[HTML]{E3DCDC}{\ul 99.57}         & \cellcolor[HTML]{E3DCDC}\textbf{79.51}  \\ \midrule[1.1pt]
Xception~\cite{chollet2017xception}                     &                      & 79.54                                  & 97.02                                  \\
RFM~\cite{wang2021representative}                          &                      & 81.34                                  & 98.26                                  \\
Add-Net~\cite{zi2020wilddeepfake}                      &                      & 72.82                                  & 97.56                                  \\
Freq-SCL~\cite{li2021frequency}                     &                      & 75.90                                  & 98.37                                  \\
MultiAtt~\cite{zhao2021multi}                     &                      & 82.33                                  & 98.82                                  \\
RECCE~\cite{cao2022end}                        &                      & {\ul 82.39}                                  & {\ul 98.82}                                 \\
\cellcolor[HTML]{E3DCDC}Ours & \multirow{-7}{*}{FS} & \cellcolor[HTML]{E3DCDC}\textbf{88.57} & \cellcolor[HTML]{E3DCDC}\textbf{99.04} \\ \bottomrule[1.1pt]
\end{tabular}
\label{tab:cross-manipulation-DF-FS}
\end{minipage}
\end{table*}

\noindent \textbf{Implementation Details.}
We tabulate configurations of the used pre-trained ViT and proposed DeepFake-Adapter in Table~\ref{tab:config}. We adopt ViT-Base~\cite{dosovitskiy2020image} pre-trained on ImageNet-21K as our frozen backbone in this paper, which is equipped with 12 blocks. These blocks are evenly split into 3 stages and thus there exist 4 blocks of ViT in each stage. In each block, every MHSA layer has 12 heads and the embedding size of every MLP layer is 3072. Since each MLP layer of ViT is paralleled with a GBA, 12 GBA are inserted in total, where the embedding size of bottleneck is 64. Moreover, we place one LSA in each stage of our network and thus the total number of LSA is 3, where each MHCA has 6 heads. In all, parameter numbers of ViT, GBA and LSA are 85.8M, 1.19M, and 15.73M. This implies the trainable dual-level adapter is much smaller than pre-trained ViT (only \textbf{19.72\%} of pre-trained ViT parameters), which statistically validates the proposed model is parameter-efficient.

All of our experiments are performed on 4 NVIDIA V100 GPUs with PyTorch framework~\cite{paszke2017automatic}. For the training schedule, we employ a 10-epochs warm-up strategy. The initial learning rate is set as 1$e-1$, with a cosine learning rate decay. We use the SGD momentum optimizer with weight decay set as $1e-4$. The batch size is set as 64.

We also provide the structure details of LSA-H (as shown in Fig.~\ref{fig:submodule} (a)) in Table~\ref{tab:LSA-H}. Specifically, each convolutional layer in blocks of Base Network, Conv-1, Conv-2, and Conv-3 is followed by a batch normalization layer and a ReLU activation function.

\noindent \textbf{Evaluation Metrics.}
We evaluate the proposed method and other baselines using the most commonly used metrics in related works~\cite{cao2022end,dong2022protecting,chen2022self,li2020face,chen2021local,qian2020thinking,zhao2021multi}, including Accuracy (ACC), Area Under the Receiver Operating Characteristic Curve (AUC), and Equal Error Rate (EER).

\begin{table}[t]
\footnotesize
\centering
\caption{Performance of cross-manipulation evaluation trained with F2F.}
\begin{tabular}{c|cccl|l}
\toprule[1.1pt]
Methods                      & Train                 & DF                            & FS                                     & NT                            & Avg.                          \\ \hline
Xception~\cite{chollet2017xception}                     &                       & 72.93                         & 64.26                                  & 70.48                         & 69.22                         \\
RFM~\cite{wang2021representative}                          &                       & 67.80                         & 64.67                                  & 64.55                         & 65.67                         \\
Add-Net~\cite{zi2020wilddeepfake}                      &                       & 70.24                         & 59.54                                  & 69.74                         & 66.51                         \\
Freq-SCL~\cite{li2021frequency}                      &                       & 67.55                         & 55.35                                  & 66.66                         & 63.19                         \\
MultiAtt~\cite{zhao2021multi}                     &                       & 73.04                         & 65.10                                  & 71.88                         & 70.01                         \\
RECCE~\cite{cao2022end}                        &                       & \textbf{75.99}                & 64.53                                  & {\ul 72.32}                         & {\ul 70.95}                \\
\cellcolor[HTML]{E3DCDC}Ours & \multirow{-7}{*}{F2F} & \cellcolor[HTML]{E3DCDC}72.24 & \cellcolor[HTML]{E3DCDC}{\ul 67.26} & \cellcolor[HTML]{E3DCDC}71.37 & \cellcolor[HTML]{E3DCDC}70.29 \\

\cellcolor[HTML]{E3DCDC}\textcolor{black}{Ours$^*$} & & \cellcolor[HTML]{E3DCDC}\textcolor{black}{{\ul 73.30}}
& \cellcolor[HTML]{E3DCDC}\textcolor{black}{\textbf{67.73}}            
& \cellcolor[HTML]{E3DCDC}\textcolor{black}{\textbf{72.39}}         
& \cellcolor[HTML]{E3DCDC}\textcolor{black}{\textbf{71.14}} \\
\bottomrule[1.1pt]
\end{tabular}
\label{tab:cross-manipulation-F2F}
\end{table}

In this section, to examine the discrimination ability of the proposed method, we carry out an intra-dataset evaluation where training and test data are from the same FF++ dataset. Following ~\cite{gu2022delving}, we compare the proposed method with a few state-of-the-art (SOTA) approaches applied in deepfake detection.
The evaluation is performed on both c23 (high-quality) and c40 (low-quality) data versions and we tabulate the comparison results in Table~\ref{tab:intra}. From Table~\ref{tab:intra}, it can be seen that in the easier case of evaluation on c23, some latest baselines have already achieved saturated performance over 99\% AUC when dealing with four manipulation types, especially in DF and FS forgeries. In such a case, although the proposed method (Ours) is not able to reach 100\% detection accuracy, it still obtains the second best performance compared to other baselines. Furthermore, in the harder case of c40, the proposed method substantially outperforms other baselines by 1\%-2\% AUC improvement in DF, FS and F2F respectively and obtains comparable results in NT. These experimental results demonstrate that the proposed method not only performs well in detection of high-quality forgeries but also is discriminative and robust in detecting low-quality forgeries filled with blur, compression and noise. This verifies exploiting high-level semantics improves the discrimination and robustness of forgery detection in presence of various post-processing. \noindent \textcolor{black}{To further improve the discriminative ability in intra-manipulation scenarios for all types of deepfake detection, we unfreeze the self-attention layer in the first block of ViT for training, which only increases a small number of trainable parameters (increase from 16.92M to 19.28M). We denote this version of the proposed method as Ours$^*$ in Table~\ref{tab:intra}. It can be seen from Table~\ref{tab:intra} that two versions of the proposed method can obtain the best or the second-best results in all settings. Notably, Ours$^*$ achieves the SOTA performance in 6/8 benchmarks and performs better than SIM under NT of c40. This further indicates that the proposed method can discriminate well all types of face manipulation methods, especially for the harder cases in c40 of FF++.} 

\begin{table}[t]
\footnotesize
\centering
\caption{Performance of cross-dataset evaluation.}
\begin{tabular}{c|c|cc|cc}
\toprule[1.1pt]
                             &                         & \multicolumn{2}{c|}{Celeb-DF}                                                   & \multicolumn{2}{c}{DFDC}                                                        \\ \cline{3-6} 
\multirow{-2}{*}{Methods}    & \multirow{-2}{*}{Train} & AUC $\uparrow$                                    & EER     $\downarrow$                                  & AUC  $\uparrow$                                    & EER    $\downarrow$                                  \\ \hline
Xception~\cite{chollet2017xception}                     &                         & 61.80                                  & 41.73                                  & 63.61                                  & 40.58                                  \\
RFM~\cite{wang2021representative}                          &                         & 65.63                                  & 38.54                                  & 66.01                                  & 39.05                                  \\
Add-Net~\cite{zi2020wilddeepfake}                      &                         & 65.29                                  & 38.90                                  & 64.78                                  & 40.23                                  \\
F3-Net~\cite{qian2020thinking}                       &                         & 61.51                                  & 42.03                                  & 64.60                                  & 39.84                                  \\
MultiAtt~\cite{zhao2021multi}                     &                         & 67.02                                  & 37.90                                  & 68.01                                  & 37.17                                  \\
RECCE~\cite{cao2022end}                        &                         & {\ul 68.71}                                  & {\ul 35.73 }                                 & {\ul 69.06 }                                 & {\ul 36.08 }                                 \\
\cellcolor[HTML]{E3DCDC}Ours & \multirow{-7}{*}{FF++}  & \cellcolor[HTML]{E3DCDC}\textbf{71.74} & \cellcolor[HTML]{E3DCDC}\textbf{33.98} & \cellcolor[HTML]{E3DCDC}\textbf{72.66} & \cellcolor[HTML]{E3DCDC}\textbf{32.68} \\
\bottomrule[1.1pt]
\end{tabular}
\label{tab:cross-dataset}
\end{table}

\begin{table*}
\footnotesize
\begin{minipage}{.65\linewidth}
\centering
\caption{Cross-dataset evaluation with single manipulation method training.}
\begin{tabular}{c|ccc|ccc|c}
\toprule[1.1pt]
                          & \multicolumn{3}{c|}{DF}                       & \multicolumn{3}{c|}{FS}                       &                        \\ \cline{2-7}
\multirow{-2}{*}{Methods} & DFDC          & Celeb-DF       & DF1.0         & DFDC          & Celeb-DF       & DF1.0         & \multirow{-2}{*}{Avg.} \\ \hline
Xception~\cite{chollet2017xception}                  & 65.4          & 68.1          & 61.7          & 70.8          & 60.1          & 60.5          & 64.4                   \\
Face X-ray~\cite{li2020face}                & 60.9          & 55.4          & 66.8          & 64.6          & 69.7          & 79.5          & 66.1                   \\
F3-Net~\cite{qian2020thinking}                      & 68.2          & 66.4          & 65.8          & 67.9          & 63.6          & 65.1          & 66.1                   \\
RFM~\cite{wang2021representative}                       & 75.8          & 72.3          & 71.7          & 71.4          & 59.1          & 71.4          & 70.2                   \\
SRM~\cite{luo2021generalizing}                       & 67.9          & 65.0          & 72.0          & 67.1          & 64.3          & {\ul 77.1 }         & 68.9                   \\
SLADD~\cite{chen2022self}                     & {\ul 77.2}          &{\ul  73.0}          & {\ul 74.2}          & {\ul 74.2}          & \textbf{80.0} & 69.5          & {\ul 74.6}                   \\
\rowcolor[HTML]{E3DCDC} 
Ours                      & \textbf{77.6} & \textbf{84.7} & \textbf{91.2} & \textbf{75.9} & {\ul 73.6}          & \textbf{81.8} & \textbf{80.8}          \\ \bottomrule[1.1pt]
\end{tabular}
\label{tab:cross-dataset-single}
\end{minipage}%
\begin{minipage}{0.4\linewidth}
\centering
\caption{Comparison with adaptation methods.}
\begin{tabular}{c|ccc}
\toprule[1.1pt]
Methods                      & Train                & DF                                     & FS                                     \\ \hline
Full-tuning                  &                      & {\ul 99.38}                                  & 74.25                                  \\
Linear-probing~\cite{zhuang2020comprehensive}               &                      & 91.41                                  & 67.20                                  \\
VPT~\cite{jia2022vpt}                          &                      & 99.37                                  & {\ul 75.93 }                                 \\
ViT-Adapter~\cite{chen2022vitadapter}                  &                      & 99.19                                  & 73.16                                  \\
\cellcolor[HTML]{E3DCDC}Ours & \multirow{-5}{*}{DF} & \cellcolor[HTML]{E3DCDC}\textbf{99.66} & \cellcolor[HTML]{E3DCDC}\textbf{76.85} \\ \midrule[1.1pt]
Full-tuning                  &                      & {\ul 87.89}                                  & 98.31                                  \\
Linear-probing~\cite{zhuang2020comprehensive}               &                      & 75.49                                  & 80.41                                  \\
VPT~\cite{jia2022vpt}                          &                      & 86.30                                  & 97.25                                  \\
ViT-Adapter~\cite{chen2022vitadapter}                  &                      & 87.65                                  & {\ul 98.55}                                  \\
\cellcolor[HTML]{E3DCDC}Ours & \multirow{-5}{*}{FS} & \cellcolor[HTML]{E3DCDC}\textbf{88.57} & \cellcolor[HTML]{E3DCDC}\textbf{99.04}\\ \bottomrule[1.1pt]
\end{tabular}
\label{tab:fine-tuning}
\end{minipage}
\end{table*}
\subsection{Evaluation on Generalization Ability}
\noindent \textbf{Cross-Manipulation Evaluation.}
To evaluate the generalization ability of our method on unseen forgeries, following RECCE~\cite{cao2022end}, we first perform cross-manipulation experiments by training and testing on different face manipulation methods in c40 version of FF++ dataset. We compare our method with several SOTAs in Table~\ref{tab:cross-manipulation-DF-FS}.
It can be observed from Table~\ref{tab:cross-manipulation-DF-FS} that the proposed method achieves better generalization performance on unseen face manipulation methods compared with other competitors, yielding about 5\% and 6\% AUC gains in two cross-manipulation evaluation settings. At the same time, the proposed method remains very competitive when training and testing data are from identical manipulation types. These results under c40 of FF++ validate that better generalizable forgery representations for deepfake detection can be captured by our method even facing highly post-processing scenarios. 

We further tabulate cross-manipulation evaluation with respect to F2F as shown in Table~\ref{tab:cross-manipulation-F2F}. Our model trained with F2F can also achieve second best average generalization performance when testing on unseen manipulation types, with very close average result to RECCE~\cite{cao2022end} and substantially surpassing the other SOTAs. \textcolor{black}{To be specific, the other version of the proposed method (Ours$^*$ in Table~\ref{tab:cross-manipulation-F2F}) further improves the cross-manipulation performance and can achieve the best average generalization ability compared to all the other SOTAs. The experimental results in Table~\ref{tab:intra} and Table~\ref{tab:cross-manipulation-F2F} demonstrate our model have great potential to deal with all types of face manipulation methods including face reenactment methods like F2F and NT.}


\noindent \textbf{Cross-Dataset Evaluation.}
To verify the generalization ability of our method on unseen forgeries with larger variations, we further conduct cross-dataset evaluations where training and testing data are from different deepfake datasets. Firstly, we perform a normal cross-dataset experiment, where we train deepfake detection models with data of c40 version on FF++, and test them on Celeb-DF and DFDC datasets, respectively. We tabulate the obtained performance of this experiment in Tables~\ref{tab:cross-dataset}. Table~\ref{tab:cross-dataset} shows that the proposed method exceeds all the other considered baselines by a large margin in terms of both AUC and EER metrics. In particular, our model can reach 71.74 \% and 72.66 \% AUC when tested on Celeb-DF and DFDC respectively, which surpass the SOTA method RECCE by about 3\%. 

To justify the generalization of the proposed method more comprehensively, we further perform another cross-dataset evaluation by training the model with a single type of manipulation method (\textit{e.g.,} DF and FS) in c23 of FF++ and testing it on the unseen DFDC, Celeb-DF and DF1.0 datasets, following the setting proposed in SLADD~\cite{chen2022self}. 
We tabulate experimental results in Table~\ref{tab:cross-dataset-single}. It is evident from Table~\ref{tab:cross-dataset-single} that the proposed method achieves the best performance regarding the cross-dataset task in most cases, by nontrivial margins improvement in some cases such as evaluation on DF1.0. Notably, our model is able to surpass other compared methods by more than 6\% AUC averaged across all cases. In all, DeepFake-Adapter can obtain quite promising performance in above two kinds of cross-dataset evaluations. This clearly demonstrates that regularized by generalizable high-level semantics of pre-trained ViT and adapted with global and local low-level forgeries via dual-level adapter, a better generalizable deepfake detection across different datasets can be obtained by our method than other existing deepfake detection methods.

\subsection{Experimental Analysis}


\noindent \textbf{Comparison of Different Adaptation Methods.}
To highlight the advantage of DeepFake-Adapter compared to other existing adaptation approaches, we compare it with the most widely-used tuning approaches, \textit{e.g.,} full-tuning and linear-probe~\cite{zhuang2020comprehensive}, and two latest parameter-efficient adaptation methods named Visual Prompt Tuning (VPT)~\cite{jia2022vpt} and ViT-Adapter~\cite{chen2022vitadapter} in Table~\ref{tab:fine-tuning}. As analyzed in Introduction, full-tuning a large ViT with limited deepfake data will result in over-fitting and distorting the pre-trained features. This is proved by the experimental results in Table~\ref{tab:fine-tuning}, where full-tuning all the parameters of ViT obtains lower performance than our method. Since tuning a small number of parameters (less than 20\% of all model parameters) can exceed tuning all the parameters of ViT, DeepFake-Adapter is proved to be both efficient and effective for deepfake detection. In addition, DeepFake-Adapter also substantially outperforms the other two parameter-efficient tuning methods, linear-probe and VPT. To achieve parameter-efficient tuning, only the last layer of whole network is adapted and tuned in linear-probe, while VPT simply prepends trainable tokens in the input space. The comparison with linear-probe and VPT in Table~\ref{tab:fine-tuning} further proves through organically interacting with features from different intermediate layers of the pre-trained ViT, the proposed dual-level adapter can attain more fine-grained and more sufficient fast adaptation. Furthermore, the proposed method also surpasses ViT-Adapter~\cite{chen2022vitadapter}. This is because our method constructs a more comprehensive dual-level adaptation, introducing LSA and GBA and thus guiding adaption process to be aware of both global and local forgery cues.
\begin{table}[t]
\footnotesize
\centering
\caption{Ablation study of Deepfake-Adapter.}
\begin{tabular}{cc|ccc}
\toprule[1.1pt]
\multicolumn{2}{c|}{Modules} & \multirow{2}{*}{Train} & \multirow{2}{*}{DF} & \multirow{2}{*}{FS} \\
GBA            & LSA            &                        &                     &                     \\ \hline
\XSolidBrush    & \XSolidBrush    & \multirow{4}{*}{DF}    &     91.41                &    67.20                  \\
\XSolidBrush    & \CheckmarkBold  &                        &      99.19               &   73.16                  \\
\CheckmarkBold  & \XSolidBrush    &                        &       99.33              &   75.03                  \\
\CheckmarkBold  & \CheckmarkBold  &                        & \textbf{99.57}      & \textbf{79.51}      \\ \midrule[1.1pt]
\XSolidBrush    & \XSolidBrush    & \multirow{4}{*}{FS}    &     75.49                &   80.41                   \\
\XSolidBrush    & \CheckmarkBold  &                        &    87.65                 &  98.55                   \\
\CheckmarkBold  & \XSolidBrush    &                        &   84.90                  &   98.99                  \\
\CheckmarkBold  & \CheckmarkBold  &                        & \textbf{88.57}      & \textbf{99.04}      \\ \bottomrule[1.1pt]
\end{tabular}
\label{tab:ablation study}
\end{table}

\begin{table}[t]
\footnotesize
\centering
\caption{Comparison of different pre-trained weights.}
\begin{tabular}{cc|ccc}
\toprule[1.1pt]
Methods                & Pre-train                          & Train                & DF                                     & FS                                     \\ \hline
                       & MAE~\cite{he2022masked}                                &                      & 99.52                                  & 73.12                                  \\
\multirow{-2}{*}{Ours} & \cellcolor[HTML]{E3DCDC}Supervised & \multirow{-2}{*}{DF} & \cellcolor[HTML]{E3DCDC}\textbf{99.57} & \cellcolor[HTML]{E3DCDC}\textbf{79.51} \\ \midrule[1.1pt]
                       & MAE~\cite{he2022masked}                                &                      & 83.97                                  & 98.88                                  \\
\multirow{-2}{*}{Ours} & \cellcolor[HTML]{E3DCDC}Supervised & \multirow{-2}{*}{FS} & \cellcolor[HTML]{E3DCDC}\textbf{88.57} & \cellcolor[HTML]{E3DCDC}\textbf{99.04} \\ \bottomrule[1.1pt]
\end{tabular}
\label{tab:init_weights}
\end{table}
\noindent \textbf{Ablation Study.} In this sub-section we investigate the impact of two key adapter modules in our DeepFake-Adapter, GBA and LSA, to the overall performance. The considered components and the corresponding results obtained for each case are tabulated in Table~\ref{tab:ablation study}. As evident from Table~\ref{tab:ablation study}, removing either GBA or LSA will degrade the overall performance. This validates that both global low-level features exploited by GBA and local low-level features extracted by LSA promote the discrimination and generalization of deepfake detection. These two modules complement each other to produce overall better performance. In addition, removing both of two adapter modules is equal to linear-prob of pre-trained ViT. It can be observed from Table~\ref{tab:ablation study} that adding either GBA or LSA will dramatically boost performance than linear-prob, justifying that any one of the proposed adapters is apparently more effective than linear-prob in terms of fast adaptation for deepfake detection. In other words, two dedicated adapters are critical and necessary to enable the whole model to achieve superior performance.
\begin{table}[t]
\footnotesize
\centering
\caption{Comparison of numbers of trainable parameters and inference time.}
\begin{tabular}{c|cc}
\toprule[1.1pt]
Methods  & \#Param  & Inference Time \\ \hline
MultiAtt~\cite{zhao2021multi} & 417.52 M & 22.81 ms  \\
SRM~\cite{luo2021generalizing}      & 53.36 M  & 11.25 ms  \\
Xception~\cite{chollet2017xception} & 20.81 M  & \textbf{4.46 ms}   \\
\textcolor{black}{LipForensics~\cite{haliassos2021lips}} & \textcolor{black}{35.99 M} & \textcolor{black}{12.63 ms} \\
\textcolor{black}{RealForensics~\cite{haliassos2022leveraging}} & \textcolor{black}{25.34 M} & \textcolor{black}{16.53 ms} \\
\rowcolor[HTML]{E3DCDC} Ours     & \textbf{16.92 M}  & {\ul 10.88 ms } \\\rowcolor[HTML]{E3DCDC} \textcolor{black}{Ours$^*$}     & \textcolor{black}{{\ul 19.28 M }} & \textcolor{black}{{\ul 10.88 ms}}  \\ \bottomrule[1.1pt]
\end{tabular}
\label{tab:num_para}
\end{table}

\noindent \textbf{Comparison of Different Pre-training Weights.} 
This section studies the effect of different pre-training weights for our model. We tabulate comparison results with MAE~\cite{he2022masked} pre-trained weights in Table~\ref{tab:init_weights}. Table~\ref{tab:init_weights} shows our backbone with supervised pre-training weights improves performance compared to self-supervised pre-training weights by MAE~\cite{he2022masked}. These results indicate supervised pre-training can provide better high-level semantics for deepfake detection and thus more suitable for adaptation. Besides, this experiment suggests our method can attain performance benefits by freely selecting the most suitable pre-training manner without additional training cost.
\begin{table*}[t]
\footnotesize
\centering
\caption{\textcolor{black}{Robustness to low-level corruptions.}}{
\textcolor{black}{\begin{tabular}{c|ccccccc|c}
\toprule[1.1pt]
Method              & Saturation             & Contrast             & Block             & Noise            & Blur             & Pixel           & Compress             & Avg.           \\ \midrule
Xception~\cite{chollet2017xception}     & 99.3          & 98.6          & \textbf{99.7} & 53.8          & 60.2          & 74.2          & 62.1          & 78.3          \\
CNN-aug~\cite{ann-aug}      & 99.3          & {\ul 99.1}          & 95.2          & 54.7          & 76.5          & 91.2          & 72.5          & 84.1          \\
Patch-based~\cite{Patch-based}  & 84.3          & 74.2          & {\ul 99.2}          & 50.0          & 54.4          & 56.7          & 53.4          & 67.5          \\
Face X-ray~\cite{li2020face}   & 97.6          & 88.5          & 99.1          & 49.8          & 63.8          & 88.6          & 55.2          & 77.5          \\
CNN-GRU~\cite{CNN-RGU}      & 99.0          & 98.8          & 97.9          & 47.9          & 71.5          & 86.5          & 74.5          & 82.3          \\
LipForensics~\cite{haliassos2021lips} & \textbf{99.9} & \textbf{99.6} & 87.4          & 73.8          & 96.1          & 95.6          & 95.6          & 92.5          \\ 
\textcolor{black}{RealForensics~\cite{haliassos2022leveraging}} & \textcolor{black}{{\ul 99.8}} & \textcolor{black}{\textbf{99.6}} & \textcolor{black}{98.9}          & \textcolor{black}{79.7}          & \textcolor{black}{95.3}         & \textcolor{black}{\textbf{98.4}}         & \textcolor{black}{97.6}          & \textcolor{black}{{\ul 95.6}}         \\ 
\rowcolor[HTML]{E3DCDC} 
Ours                                        & 97.6          & 97.2          & 97.4          & {\ul 86.0} & \textbf{96.9} & 95.8 & {\ul 97.6} & 95.5 \\
\rowcolor[HTML]{E3DCDC} 
\textcolor{black}{Ours$^*$}                                       & \textcolor{black}{97.9}         & \textcolor{black}{97.5}          & \textcolor{black}{98.0}         & \textcolor{black}{\textbf{86.3}} & \textcolor{black}{{\ul 96.8}} & \textcolor{black}{{\ul 96.2}} & \textcolor{black}{\textbf{98.0}} & \textcolor{black}{\textbf{95.8}} \\
\bottomrule[1.1pt]
\end{tabular}
}}
\label{tab:low-level}
\end{table*}

\noindent \textbf{Comparison of Numbers of Trainable Parameters and Inference Time.} 
We further compare the number of trainable parameters and inference time of our method with some representative baselines in Table~\ref{tab:num_para}. \textcolor{black}{The measurement of inference time is performed based on GPU: Tesla V100-PCIE-32GB with 32GB space, and CPU: Intel(R) Xeon(R) Gold 6278C CPU @ 2.60GHz.} Comparison results in Tables~\ref{tab:intra}--\ref{tab:cross-dataset-single} and Table~\ref{tab:low-level} have demonstrated our method outperforms these baselines. Table~\ref{tab:num_para} here further indicates such better performance of our method is achieved by using significantly fewer trainable parameters and less inference time (only slightly more than Xception), implying the efficacy and efficiency of the proposed DeepFake-Adapter.

\noindent \textbf{Robustness to low-level corruptions.}  We follow~\cite{haliassos2021lips} to assess robustness to various unseen low-level perturbations. Specifically, following~\cite{haliassos2021lips}, we compare the proposed method with other SOTA deepfake detection methods trained on FF++ c23 dataset on seven unseen low-level perturbations, such as saturation , contrast, block, noise, blur, pixel, and compress, as illustrated in Table~\ref{tab:low-level}. \textcolor{black}{It can be seen that the proposed method (Ours and Ours$^*$) can achieve the best and the second-best performance under most of the low-level perturbations, and Ours$^*$ attains the best average performance compared to all the other baselines.} This further implies that DeepFake-Adapter aided by high-level semantic understanding capability from large vision models is more robust to unseen low-level corruptions.

\noindent \textcolor{black}{\textbf{Detection of deepfake samples generated by diffusion modes.} We further test our model on deepfake samples generated by 5 representative diffusion modes, such as DDPM~\cite{ho2020denoising}, IDDPM~\cite{nichol2021improved}, ADM~\cite{dhariwal2021diffusion}, PNDM~\cite{liu2022pseudo}, and LDM~\cite{rombach2022high}. Specifically, our model and Xception~\cite{chollet2017xception} are trained on the FF++ c40 dataset and evaluated under the setting of~\cite{ricker2022towards}. Note that both our model and Xception are evaluated in a zero-shot testing setup. We conducted testing using the officially provided evaluation scripts. The PD metric refers to the probability of detection at a fixed false alarm rate, which is defined as the true positive rate at a specific false alarm rate. PD@10\% indicates the probability of our model detecting fake images while allowing a 10\% false alarm rate, a higher value is desirable. The same applies to PD@5\% and PD@1\%.}

\textcolor{black}{We tabulate the comparison between our model and Xception~\cite{chollet2017xception} in Table~\ref{tab:diffusion}. It can be seen that the proposed model outperforms the baseline method in detecting deepfake samples generated by all the 5 representative diffusion modes, under all evaluation settings. This suggests the proposed method can be applied and generalized well to more realistic deepfake images generated by diffusion models.}
\begin{table}[t]
\footnotesize
\centering
\caption{\textcolor{black}{Performance of detection of deepfake samples generated by various diffusion modes.}}
\resizebox{\linewidth}{!}{
\textcolor{black}{\begin{tabular}{c|c|ccccc}
\toprule[1.1pt]
DM                             & Methods                      & AP                                    & AUROC                                 & PD@10\%                               & PD@5\%                                & PD@1\%                               \\ \hline
                               & Xception~\cite{chollet2017xception}              & 51.4                                  & 50.9                                  & 11.9                                  & 6.0                                   & 1.1                                  \\
\multirow{-2}{*}{DDPM \cite{ddpm2020}}  & \cellcolor[HTML]{D0CECE}Ours & \cellcolor[HTML]{D0CECE}\textbf{59.9} & \cellcolor[HTML]{D0CECE}\textbf{61.9} & \cellcolor[HTML]{D0CECE}\textbf{18.4} & \cellcolor[HTML]{D0CECE}\textbf{10.6} & \cellcolor[HTML]{D0CECE}\textbf{2.6} \\ \hline
                               & Xception~\cite{chollet2017xception}              & 52.8                                  & 52.4                                  & 13.7                                  & 6.6                                   & 1.2                                  \\
\multirow{-2}{*}{IDDPM \cite{iddpm2021}} & \cellcolor[HTML]{D0CECE}Ours & \cellcolor[HTML]{D0CECE}\textbf{61.8} & \cellcolor[HTML]{D0CECE}\textbf{64.2} & \cellcolor[HTML]{D0CECE}\textbf{20.3} & \cellcolor[HTML]{D0CECE}\textbf{11.5} & \cellcolor[HTML]{D0CECE}\textbf{2.8} \\ \hline
                               & Xception~\cite{chollet2017xception}              & 47.9                                  & 47.7                                  & 8.3                                   & 3.7                                   & 0.5                                  \\
\multirow{-2}{*}{ADM~\cite{adm2021}}   & \cellcolor[HTML]{D0CECE}Ours & \cellcolor[HTML]{D0CECE}\textbf{54.5} & \cellcolor[HTML]{D0CECE}\textbf{56.9} & \cellcolor[HTML]{D0CECE}\textbf{13.2} & \cellcolor[HTML]{D0CECE}\textbf{6.6}  & \cellcolor[HTML]{D0CECE}\textbf{1.1} \\ \hline
                               & Xception~\cite{chollet2017xception}             & 48.7                                  & 46.8                                  & 10.6                                  & 5.4                                   & 1.1                                  \\
\multirow{-2}{*}{PNDM~\cite{pndm2022}}  & \cellcolor[HTML]{D0CECE}Ours & \cellcolor[HTML]{D0CECE}\textbf{53.4} & \cellcolor[HTML]{D0CECE}\textbf{54.8} & \cellcolor[HTML]{D0CECE}\textbf{12.7} & \cellcolor[HTML]{D0CECE}\textbf{6.8}  & \cellcolor[HTML]{D0CECE}\textbf{1.3} \\ \hline
                               & Xception~\cite{chollet2017xception}              & 49.8                                  & 49.3                                  & 10.4                                  & 5.1                                   & 1.0                                  \\
\multirow{-2}{*}{LDM~\cite{ldm2022}}   & \cellcolor[HTML]{D0CECE}Ours & \cellcolor[HTML]{D0CECE}\textbf{57.5} & \cellcolor[HTML]{D0CECE}\textbf{59.4} & \cellcolor[HTML]{D0CECE}\textbf{16.3} & \cellcolor[HTML]{D0CECE}\textbf{8.9}  & \cellcolor[HTML]{D0CECE}\textbf{1.8} \\ \bottomrule[1.1pt]
\end{tabular}}}
\label{tab:diffusion}
\end{table}

\noindent \textcolor{black}{\textbf{Comparison of Different ViT Architectures.} This section studies the proposed DeepFake-Adapter with various ViT architectures, in the cross-manipulation evaluation setting same as in Section 4.3. Particularly, we tabulate the cross-manipulation performance of DeepFake-Adapter based on ViT-Base and ViT-Large architectures in Table~\ref{tab:ViTcomp}. It shows that DeepFake-Adapter based on various ViT architectures can obtain the best or second-best performance compared to the SOTA method RECCE in both intra- and cross-manipulation settings. This suggests the proposed DeepFake-Adapter with various ViT architectures can simultaneously yield promising performance. Meanwhile, the t-SNE plots of different ViT architectures' features are displayed in Figure~\ref{fig:t-sne}. All these qualitative and quantitative results validate that the proposed DeepFake-Adapter can be compatible with various ViT architectures for the task of deepfake detection.}

\noindent \textcolor{black}{\textbf{Comparison between Different Numbers of Stages and Blocks.} This section studies the choice regarding the number of Stages and Blocks. As mentioned above, the proposed DeepFake-Adapter splits the pre-trained ViT into 3 Stages with 4 Blocks in each Stage. This choice is determined by the consideration of both effectiveness and efficiency. The core claim of the proposed DeepFake-Adapter is a parameter-efficient tuning approach for deepfake detection. As shown in Table~\ref{tab:config}, LSA possesses much more trainable parameters than GBA. Therefore, employing LSA on every block will significantly increase the trainable parameters and thus make the proposed method less parameter-efficient. For this reason, we introduce the concept of Stage and deploy only one LSA in each Stage, which saves many trainable parameters. This is proved by the fact that DeepFake-Adapter outperforms the full-tuning adaptation method by only tuning less than 20\% of all model parameters. On the other hand, decreasing the number of GBA and LSA would also affect the performance.}

\textcolor{black}{To validate the efficacy of this choice, we tabulate the comparison between different number of Stages and Blocks in Table~\ref{tab:cmpstages}. It follows the setting of cross-manipulation evaluation in Sec. 4.3. As illustrated in Table~\ref{tab:cmpstages}, the setting of 6 Stages with 2 Blocks dramatically increases the number of trainable parameters but still achieves less effective performance. Meanwhile, the setting of 2 Stages with 6 Blocks yields fewer trainable parameters at the cost of degraded performance. In contrast, the setting (3 Stages with 4 Blocks) adopted by DeepFake-Adapter can simultaneously achieve moderate number of trainable parameters and best performance. This indicates the more optimal choice regarding the number of Stages and Blocks by the proposed method.}

\subsection{Visualization}

\begin{figure}[t] 
	\begin{center}
		\includegraphics[width=1\linewidth]{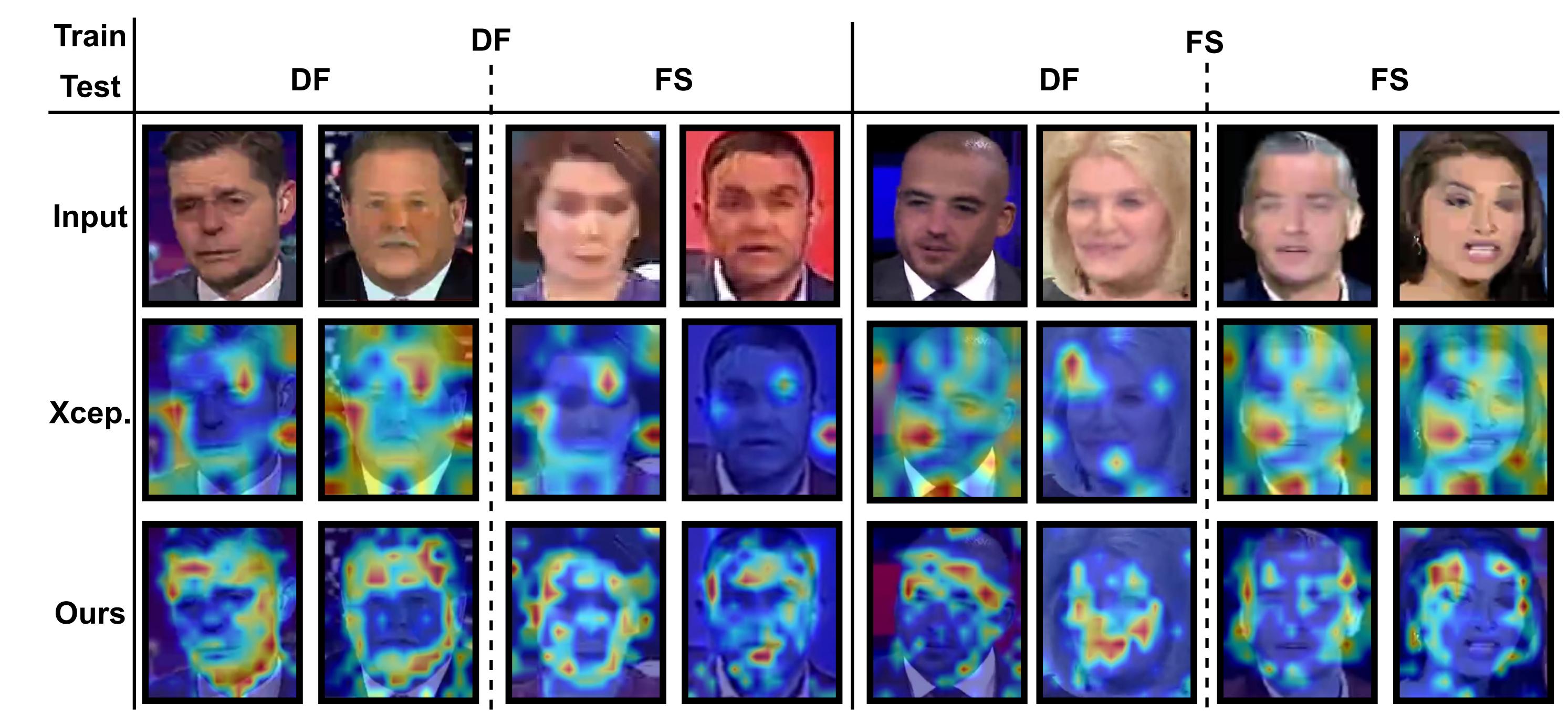}
	\end{center}
	\caption{Comparison of Grad-CAM visualizations between Xception and the proposed model in cross-manipulation evaluation. (Best viewed in color)}
	\label{fig:gradcam-cross-manipulation}
\end{figure}

\begin{figure*}[t] 
	\begin{center}
		\includegraphics[width=1\linewidth]{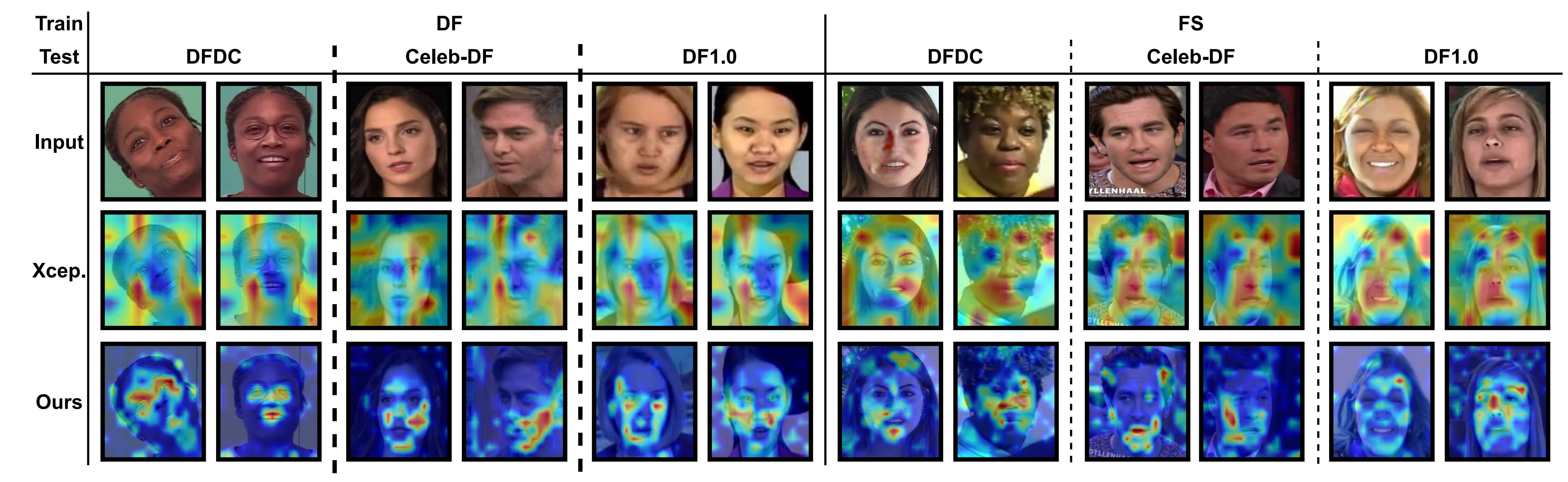}
	\end{center}
	\caption{Comparison of Grad-CAM visualizations between Xception and the proposed model in cross-dataset evaluation among DFDC, Celeb-DF and DF1.0 datasets. (Best viewed in color)}
	\label{fig:gradcam-cross-dataset}
\end{figure*}

\begin{figure}[t] 
	\begin{center}
		\includegraphics[width=1\linewidth]{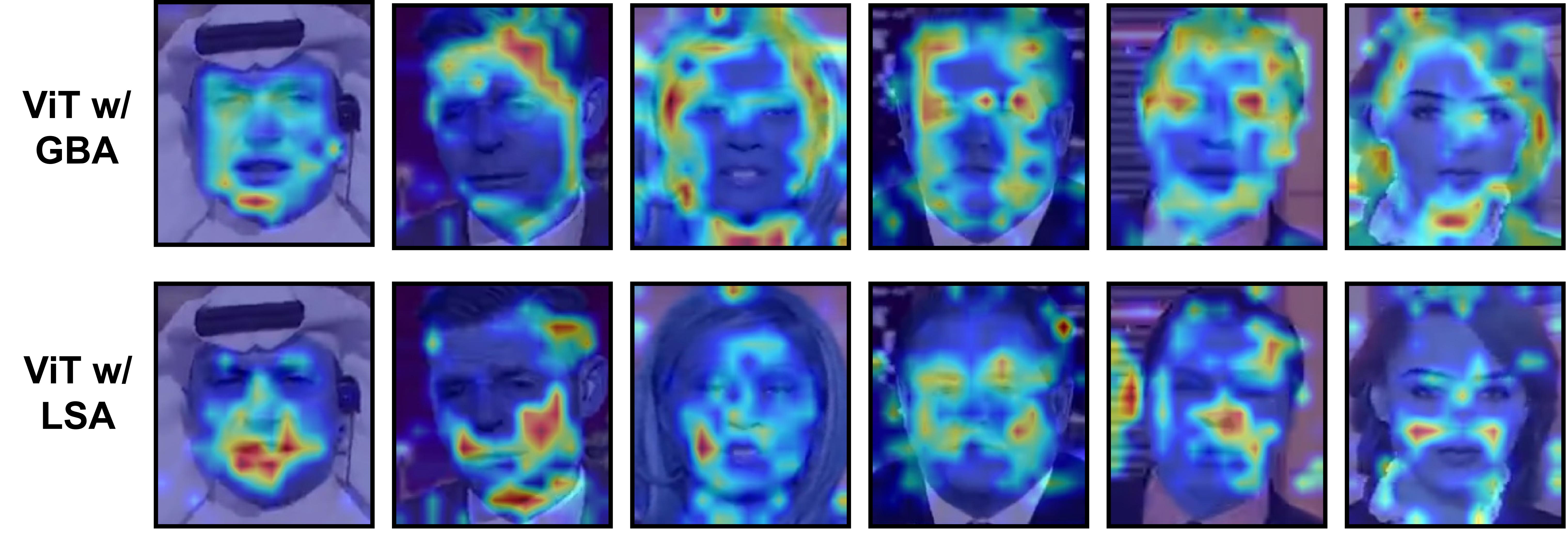}
	\end{center}
	\caption{Comparison of Grad-CAM visualizations between ViT backbone with GBA and LSA. (Best viewed in color)}
	\label{fig:gradcam_GBA_LSA}
\end{figure}

\begin{figure}[t] 
	\begin{center}
		\includegraphics[width=1\linewidth]{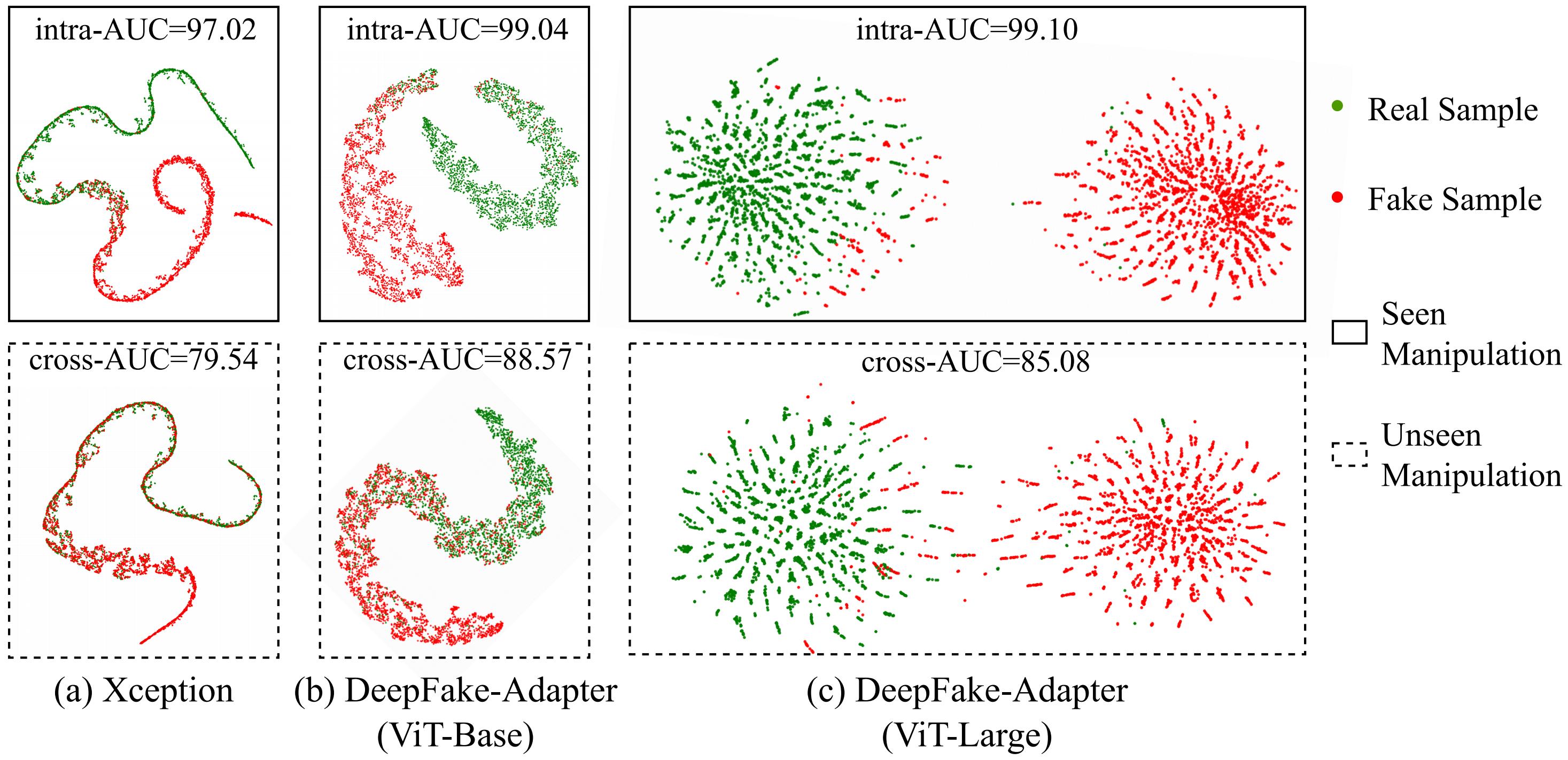}
	\end{center}
	\caption{\textcolor{black}{t-SNE visualization of features encoded by (a) Xception (b) DeepFake-Adapter (ViT-Base) and (c) DeepFake-Adapter (ViT-Large) in intra and cross-manipulation settings. (Best viewed in color)}}
	\label{fig:t-sne}
\end{figure}
\noindent \textbf{Attention map compared with Xception.}
To provide a deeper understanding about the decision-making mechanism of our method, we compare Grad-CAM~\cite{selvaraju2017grad} visualizations between our model and Xception~\cite{chollet2017xception} on FF++ as shown in Fig.~\ref{fig:gradcam-cross-manipulation}. Some critical observations can be derived from Fig.~\ref{fig:gradcam-cross-manipulation} that \textbf{1)} Xception sometimes pays attention to semantically irrelevant regions for real/fake classification such as background, especially when it is trained with DF and tested on unseen manipulation FS. In contrast, in both intra and cross-manipulation evaluations, our model is more likely to focus on facial regions or facial contours. Since face swapping is bound to leave some manipulation traces in local textures of facial regions or blending boundaries around facial contours, these visualizations demonstrate that our model is able to locate such generic regions to exploit generalizable forgery patterns and \textbf{2)} our model can generate more fine-grained and adaptive attention heatmaps than Xception. This indicates more sufficient discrimination cues can be captured by the proposed method. 

We further show more samples in the cross-dataset scenario as illustrated in Fig.~\ref{fig:gradcam-cross-dataset}. As illustrated in Fig.~\ref{fig:gradcam-cross-dataset}, Xception usually pays attention to background or random large regions irrelevant for face forgery detection, while the proposed model adaptively detects more fine-grained forgery patterns in the facial regions. These visualizations further demonstrate the generalization ability of our model for deepfake detection.

\noindent \textbf{Attention map regarding GBA and LSA.}
To facilitate the understanding about the function of GBA and LSA, we integrate the pre-trained ViT only with GBA or LSA and visualize Grad-CAM~\cite{selvaraju2017grad} of the two cases in Fig.~\ref{fig:gradcam_GBA_LSA}, respectively. It can be observed from Fig.~\ref{fig:gradcam_GBA_LSA} that \textbf{1)} ViT adapted by GBA tends to focus on facial contours to exploit global low-level features related to the blending boundary, while \textbf{2)} ViT adapted with LSA pays attention to some facial regions to adaptively capture local low-level features about textures. This means the adaption by GBA and LSA could complement each other and the combination of them would contribute to a more sufficient and comprehensive adaption. 

\begin{table}[t]
\footnotesize
\centering
\caption{\textcolor{black}{Performance of cross-manipulation evaluation based on various ViT architectures.}}
\textcolor{black}{\begin{tabular}{c|ccc}
\toprule[1.1pt]
Methods          & Train               & DF             & FS             \\ \hline
RECCE~\cite{cao2022end}    & \multirow{3}{*}{DF} & \textbf{99.65} & 74.29          \\
Ours (ViT-Base)  &                     & 99.57          & \textbf{79.51} \\
Ours (ViT-Large) &                     & {\ul 99.62}    & {\ul 78.60}    \\ \midrule[1.1pt]
RECCE~\cite{cao2022end}    & \multirow{3}{*}{FS} & 82.39          & 98.82          \\
Ours (ViT-Base)  &                     & \textbf{88.57} & {\ul 99.04}    \\
Ours (ViT-Large) &                     & {\ul 85.08}    & \textbf{99.10} \\ \bottomrule[1.1pt]
\end{tabular}}
\label{tab:ViTcomp}
\end{table}

\begin{table}[t]
\footnotesize
\centering
\caption{\textcolor{black}{Comparison between different numbers of Stages and Blocks in each Stage. Training data is DF of FF++ dataset and testing data is DF and FS of FF++ dataset.}}
\textcolor{black}{\begin{tabular}{c|ccc}
\toprule[1.1pt]
\#Stages and Blocks    & \#Param & DF             & FS             \\ \hline
2 Stages with 6 Blocks & 12.19 M & 99.34          & 75.44          \\
6 Stages with 2 Blocks & 31.11 M & {\ul 99.53}    & {\ul 78.69}    \\
\rowcolor[HTML]{D0CECE} 3 Stages with 4 Blocks & 16.56 M & \textbf{99.57} & \textbf{79.51} \\ \bottomrule[1.1pt]
\end{tabular}}
\label{tab:cmpstages}
\end{table}

\noindent \textbf{Feature Distribution.} We apply t-SNE~\cite{van2008visualizing} to visualize feature embeddings under the cross-manipulation setting, as illustrated in Fig.~\ref{fig:t-sne}. \textcolor{black}{Plots in solid-lined boxes, such as plots in the first row of Fig.~\ref{fig:t-sne}, visualize features of testing samples from seen manipulations encoded by Xception and DeepFake-Adapter (based on ViT-Base and ViT-Large architectures), respectively. Meanwhile, plots in dotted-lined boxes, such as plots in the second row of Fig.~\ref{fig:t-sne}, visualize features of testing samples from unseen manipulations encoded by Xception and DeepFake-Adapter, respectively. }

\textcolor{black}{As illustrated in the first row of Fig.~\ref{fig:t-sne}, when training and testing manipulation types are identical, both Xception and our model can attain clear classification boundary with two separated class clusters. However, the second row of Fig.~\ref{fig:t-sne} shows that feature embeddings of Xception become heavily overlapped between real/fake samples once facing unseen manipulations, while our method can remain much clearer clusters with smaller overlaps. This demonstrates that our model can learn superior generalizable representations for deepfake detection.}

\textcolor{black}{Furthermore, it can be seen from (b) and (c) of Fig.~\ref{fig:t-sne} that our model based on both ViT-Base and ViT-Large architectures can achieve clear classification boundaries with two separated class clusters. This demonstrates the proposed DeepFake-Adapter can be compatible with various ViT architectures for the task of deepfake detection.}

\section{Conclusion}
This paper studies high-level semantics for deepfake detection and first introduces the adapter approach to efficiently tune large pre-trained ViT to our task. A powerful DeepFake-Adapter is devised with GBA and LSA, which effectively and efficiently leads high-level semantics of ViT to interact with global and local low-level features in a dual-level fashion. Various quantitative and qualitative experiments demonstrate the effectiveness of our model for deepfake detection. Valuable observations pave the way for future research on generalizable deepfake detection in the era of large vision models.

\noindent \textbf{Potential Negative Impact.} Although some face forgery data is used, this work is designed to help people better fight against the abuse of deepfake technology. Through our study and releasing our code, we hope to draw greater attention towards generalizable deepfake detection.

\section{acknowledgements}
This study is supported by National Natural Science Foundation of China (Grant No. 62306090 and Grant No. 62236003); Natural Science Foundation of Guangdong Province of China (Grant No. 2024A1515010147); Shenzhen College Stability Support Plan (Grant No. GXWD20220817144428005). This study is supported by the Ministry of Education, Singapore, under its MOE AcRF Tier 2 (MOET2EP20221-0012), NTU NAP, and under the RIE2020 Industry Alignment Fund – Industry Collaboration Projects (IAF-ICP) Funding Initiative, as well as cash and in-kind contribution from the industry partner(s).

\section{Data Availability Statement}
The datasets analysed during this study are all publicly available for the research purpose - the \href{https://github.com/ondyari/FaceForensics}{FaceForensics++}, \href{https://github.com/yuezunli/celeb-deepfakeforensics}{Celeb-DF}, \href{https://www.kaggle.com/c/deepfake-detection-challenge}{Deepfake Detection Challenge}, and \href{https://github.com/EndlessSora/DeeperForensics-1.0}{DeeperForensics-1.0} datasets.

%
%

\bibliographystyle{spmpsci}      
\bibliography{ref}   


\end{document}